\documentclass[letterpaper,10pt,final,left=0.75in,right=0.75in,top=1in,bottom=1in,onecolumn,nocentering]{IEEEtran}

\IEEEoverridecommandlockouts

\usepackage{graphicx}
\usepackage{amsfonts}
\usepackage{amsmath}
\usepackage{amssymb}
\usepackage{color}
\usepackage{cite}
\usepackage{graphicx,color,overpic}
\usepackage{psfrag}
\usepackage{amssymb}
\usepackage{times}
\usepackage{latexsym}
\usepackage{bm}
\usepackage{cases}
\usepackage{array}
\usepackage{fancyhdr}
\usepackage{setspace}
\usepackage{subfigure}
\usepackage{url}
\usepackage{multirow}
\usepackage{epstopdf}

\usepackage{epsfig}
\usepackage{fancybox}
\usepackage{textcomp}

\usepackage{lipsum}

\newcommand{\SIM}[0]{\mathrm{sim}}
\newcommand{\x}[0]{\mathbf{x}}
\newcommand{\z}[0]{\mathbf{z}}
\newcommand{\F}[0]{\mathbf{F}}
\newcommand{\f}[0]{\mathbf{f}}

\newcommand{\s}[0]{\mathbf{s}}
\newcommand{\C}[0]{\mathcal{C}}

\IEEEoverridecommandlockouts                              % This command is only needed if 
\usepackage{times} % assumes new font selection scheme installed

\title{ {\fontsize{24}{24}\selectfont \bf Crowd-sensing Simultaneous Localization and Radio Fingerprint Mapping based on Probabilistic Similarity Models}
}

\author{Ran Liu, Sumudu Hasala Marakkalage, Madhushanka Padmal, Thiruketheeswaran Shaganan, \\
Chau Yuen, Yong Liang Guan, and U-Xuan Tan
\thanks{This work is partially supported by the National Science Foundation of China (No. 61601381, 61750110529, and 61701421) and the Sichuan Science and Technology Program (No. 2019YFH0161).} 
\thanks{R. Liu, S. H. Marakkalage, C. Yuen, and U-X. Tan are with the Engineering Product Development Pillar, Singapore University of Technology and Design, 8 Somapah Rd, Singapore, 487372 
{\{\tt\small ran\_liu, yuenchau, uxuan\_tan\}@sutd.edu.sg}.
}
\thanks{
Y. L. Guan is with the School of Electrical and Electronics Engineering, Nanyang Technological University, 50 Nanyang Avenue, Singapore 639798 
{\tt\small eylguan@ntu.edu.sg}. 
}
\thanks{M. Padmal and T. Shaganan are with the Department of Electronic and Telecommunication Engineering, University of Moratuwa, Sri Lanka, 10400.
}
}

\begin{document}

\maketitle
\thispagestyle{empty}
\pagestyle{empty}

%\begin{IEEEbiography}{Ran Liu}
%\end{IEEEbiography}

%\begin{IEEEbiographynophoto}{Chau~Yuen}

%\newpage

%%%%%%%%%%%%%%%%%%%%%%%%%%%%%%%%%%%%%%%%%%%%%%%%%%%%%%%%%%%%%%%%%%%%%%%%%%%%%%%%
\begin{IEEEkeywords}
\textbf{Ran Liu} received the bachelor's degree from the Southwest University of Science and Technology, Mianyang, China, in 2007, 
and the Ph.D. degree from the University of Tuebingen, Germany, in 2014, under the supervision of Prof. Dr. A. Zell and Prof. Dr. A. Schilling. 
Since 2014, he has been a Post-Doctoral Research Fellow under the supervision of Prof. C. Yuen with the MIT International Design Center, 
Singapore University of Technology and Design. 
His research interests include robotics, indoor positioning, UHF RFID localization, and mapping.

\textbf{Chau Yuen} (S'02-M'08-SM'12) received the B.Eng. and Ph.D. degrees from
Nanyang Technological University, Singapore, in 2000 and 2004, respectively.
He was a Post-Doctoral Fellow with Lucent Technologies Bell Labs, Murray Hill, in 2005. From 2006 to 2010, he was with the Institute for Infocomm Research, Singapore, as a Senior Research Engineer. He was a Visiting
Assistant Professor with The Hong Kong Polytechnic University in 2008. He has been with the Singapore University of Technology and Design since 2010.
In 2012, he received the IEEE Asia-Pacific Outstanding Young Researcher Award. 
He serves as an Editor of the IEEE TRANSACTIONS ON COMMUNICATIONS and the IEEE TRANSACTIONS ON VEHICULAR TECHNOLOGY.

\textbf{Yong Liang Guan} received the Ph.D. degree from the Imperial College of Science, Technology and Medicine, University of London, in 1997, 
and the B.Eng. degree (Hons.) from the National University of Singapore in 1991. 
He is currently an Associate Professor with the School of Electrical and Electronic Engineering, Nanyang Technological University. 
His research interests include modulation, coding and signal processing for communication, information security and storage systems.

\textbf{U-Xuan Tan} (S'07-M'10) received the B.Eng. and Ph.D. degrees from Nanyang Technological University, Singapore, in 2005 and 2010, respectively.
From 2009 to 2011, he was a Post-Doctoral Fellow with the University of Maryland, College Park, MD, USA. 
From 2012 to 2014, he was a Lecturer with the Singapore University of Technology and Design. 
In 2014, he decided to take up a research intensive role and has since been an Assistant Professor with the Singapore University of Technology and Design. 
His research interests include mechatronics, medical robotics, sensing, control, mechanism design, and disturbance compensation. 
He received the STUD Outstanding Education Award-Excellence in Teaching in 2015 and the Best Student Paper Finalist for the IEEE ROBIO 2006.
\end{IEEEkeywords}
%%%%%%%%%%%%%%%%%%%%%%%%%%%%%%%%%%%%%%%%%%%%%%%%%%%%%%%%%%%%%%%%%%%%%%%%%%%%%%%%

%%%%%%%%%%%%%%%%%%%%%%%%%%%%%%%%%%%%%%%%%%%%%%%%%%%%%%%%%%%%%%%%%%%%%%%%%%%%%%%%
\begin{abstract}
Simultaneous localization and mapping (SLAM) has been richly researched in past years particularly with regard to range-based or visual-based sensors.
%The wide deployment of ubiquitous wireless network offers an opportunity for localization and mapping using the radio signals. 
Instead of deploying dedicated devices that use visual features, 
it is more pragmatic to exploit the radio features to achieve this task, 
due to their ubiquitous nature and the wide deployment of Wifi wireless network.
%which has in-built sensing capabilities (wireless network connectivity, accelerometer, gyro etc.), 
%to exploit the radio features of the environment.
%achieve this task, 
%due to their ubiquitous nature. 
In this paper, 
we present a novel approach for crowd-sensing simultaneous localization and radio fingerprint mapping (C-SLAM-RF) in large unknown indoor environments. 
The proposed system makes use of the received signal strength (RSS) from surrounding Wifi access points (AP) 
and the motion tracking data from a smart phone (Tango as an example). 
These measurements are captured duration the walking of multiple users in unknown environments without map information and location of the AP.
The experiments were done in a university building with dynamic environment and the results show that the proposed system is capable of estimating the tracks of a group of users with an accuracy of 1.74 meters when compared to the ground truth acquired from a point cloud-based SLAM. 
%The current signal-strength-based SLAM system has to rely on the modeling of the of radio signal propagation, 
%which is not applicable in harsh indoor environment.
%The error, although, is larger as compared to the laser-based positioning system used in robotics, we believe the accuracy we achieved so far is much better than the radio fingerprinting-based approach. 
\end{abstract}
%%%%%%%%%%%%%%%%%%%%%%%%%%%%%%%%%%%%%%%%%%%%%%%%%%%%%%%%%%%%%%%%%%%%%%%%%%%%%%%%

%%%%%%%%%%%%%%%%%%%%%%%%%%%%%%%%%%%%%%%%%%%%%%%%%%%%%%%%%%%%%%%%%%%%%%%%%%%%%%%%%%%%%%%%%%%%%%%%%%%%%%%%
\section{\textbf{Introduction}}
\label{Introduction}
%%%%%%%%%%%%%%%%%%%%%%%%%%%%%%%%%%%%%%%%%%%%%%%%%%%%%%%%%%%%%%%%%%%%%%%%%%%%%%%%%%%%%%%%%%%%%%%%%%%%%%%%

Over the past few decades, researchers are working on developing efficient methods and technologies to map the unknown environment 
and localize mobile devices (robots and smartphones) 
in that environment \cite{dissanayake2001solution, montemerlo2002fastslam, ferris2007wifi}. 
This process is well known by the term, Simultaneous Localization and Mapping (SLAM). 
Extensive researches have been done with regard to visual-based or range-based SLAM. 
It is elementary to detect loop closures, either implicitly 
(i.e., the extended Kalman filter-based \cite{montemerlo2002fastslam} or the particle filter-based SLAM \cite{Thrun_Probabilistic_robotics}) or explicitly (i.e., the graph-based SLAM \cite{kuemmerle11icra}), 
that permits to correct the accumulated odometry error. 

%Loop closure denotes a situation that the user re-visits a previous observed place. 
In order to perform loop closure detection in SLAM, 
dedicated device (i.e., laser range finder or visual camera) is required to measure the similarity of observations by 
scan matching \cite{Lu_millos_1997} or feature matching \cite{Taketomi2017, lahiru_access}, which are usually computationally expensive. 
%which usually requires high computational demand. 
However, growing popularity of Wifi wireless network provides a new opportunity to detect loop closure and perform SLAM in a different way. 

%Such a need emerged because of the following reason.
%In outdoor environments, Global Positioning Systems (GPS) are widely adopted for mapping and localizing because of their high accuracy. However, GPS cannot be exploited for indoor environment mapping, due to poor GPS coverage in such environments. 
%Hence, it is paramount to have an alternative method to map indoor environments with better accuracy.

%The wide deployment of wireless network provides an opportunity to localize and mapping using the radio signals. 
Most existing building with Wifi network deployed 
can be exploited for localization and mapping with low hardware requirement and computational cost due to their ubiquitous nature of in-built sensing capabilities \cite{ran_ieee_sensors2017, billy_iot, ran_localize_AP, liuran_spawc}. 
%Instead of developing dedicated hardware devices, it is more pragmatic to utilize mobile devices with in-built sensing capabilities (wireless network connectivity, accelerometer, gyro etc.), 
%to achieve the aforementioned task, due to their ubiquitous nature. 
%Existing WiFi indoor localization methods generally fall into two main categories such as, 
%model-based \cite{yang2012locating} or fingerprinting-based. 
%Model-based techniques determine locations based on geometric models \cite{grisettiyz2005improving,jensen2009graph,dellaert1999monte}. 
The current signal-strength-based SLAM system has 
to use an analytical model to feature the radio signal propagation \cite{ferris2007wifi, Huang_wifi_slam_11}. 
However, it is not practical to build such a model due to 
multiple path propagation issue in indoor environments with unexpected occlusions. 
%However, the focus of this paper is on fingerprint-based methods due to wide availability of wireless networks in crowded urban indoor environments such as office buildings, shopping malls, airports etc. Moreover, fingerprint-based methods are easy to implement when compared to LiDAR (expensive) \cite{biswas2010wifi}, and camera (require more user involvement) \cite{mulloni2009indoor}. 
Radio frequency (RF) fingerprinting \cite{Yassin_ieee_tutorials_2016, he2016wi}, on the other hand, represents a location with a set of radio signals, which is considered to be more robust against the signal distortions. 
Therefore, we adopt this technique to simultaneously localize a user and create a radio map of the environment.

%Instead of using physical feature (e.g. wall, door) of a building in building a map, 
%we use the RF fingerprint to build a virtual map.. and use that for the localization.. 
%The resulted fingerprinting map can be used for the positioning of other users afterwards, 
In opposite to the feature map or occupancy map built by visual cameras or laser range finders, 
our goal is to build a map (in particular a radio map) with RF fingerprint as feature, and use that for the localization.
%A fingerprint database has to be generated by site surveys which require intensive cost on manpower and time. 
%In this case, the location of a user is determined by the most similar fingerprint in the radio map. 
To ensure a good positioning accuracy, 
it is essential to have a fine-grained radio map \cite{Yassin_ieee_tutorials_2016, he2016wi} and the one created by a single user obviously cannot satisfy the requirement.
%Additionally, the dynamic nature of environments require to update the database time to time, which is not cost effective. 
Therefore, a low cost method (e.g., acquire fingerprints via  crowd-sensing by multiple users) to create the radio map is a necessity.

%Model-based techniques determine locations based on geometric models \cite{grisettiyz2005improving,jensen2009graph,dellaert1999monte}. 
%However, the focus of this paper is on fingerprint-based methods due to wide availability of wireless networks in crowded urban indoor environments such as office buildings, shopping malls, airports etc. Moreover, fingerprint-based methods are easy to implement when compared to LiDAR (expensive) \cite{biswas2010wifi}, and camera (require more user involvement) \cite{mulloni2009indoor}. 

In this paper, 
we propose to fuse the motion tracking data from a pedestrian dead reckoning system 
and received signal strength (RSS) measurements from surrounding Wifi access point (AP), 
to estimate the trajectory of the users and map the radio fingerprints in unknown environments via a crowd-sensing fashion, 
using graph SLAM technique in the system back-end. 
%Although the dedicated Tango phone (Lenovo Phab2 Pro) is used to perform pedestrian dead reckoning, 
%it can be replaced by other low-cost and pervasive tracking units 
Although our inertial tracking unit is Tango, our system can be extended 
to include any devices that provide inertial tracking functions (for example low-cost IMU sensors in a mobile phone). 
We adopt the crowd-sensing nature in our system to conveniently map large indoor environments using multiple mobile devices. 
%Google Tango API provides an accurate motion tracking framework when compared to that of non Tango supported phones. 
%It can provide the position and orientation (pose) of the device in full six degrees of freedom (6-DOF) \cite{GoogleTango}. 
The proposed approach 
corrects the trajectory of multiple users by the radio fingerprints taken during the exploration of the environment without any prior knowledge of the infrastructure.
%generate the WiFi fingerprint map without any site survey. 
Our system leverages the in-built sensing capabilities of smartphones and 
the crowd-sensing nature enables to further generate the fingerprint map in larger environments at lower cost, in contrast to traditional site surveying methods.

The key contributions of this paper are:
\begin{itemize}
	\item We propose to incorporate Wifi fingerprint and motion information for crowd-sensing SLAM in an unknown indoor environment;
	\item We propose an algorithm that automatically learns a model to characterize the uncertainty of a loop based on the degree of similarity using the short term odometry data;
       \item  We thoroughly evaluate our approach in one building at our campus with an area of approx. 9000 square meters.
\end{itemize}

We organize the rest of this paper as follows. 
%Section \ref{related_work} discusses related work on SLAM localization methods with WiFi fingerprinting and motion tracking.
We discuss the related work in Section \ref{related_work}. 
In Section \ref{system_overview}, we give an overview of the system and explain each component afterwards. 
In section \ref{experimental_evaluations}, we present the results of our experiments done using the proposed method. 
Finally, Section \ref{conclusions} concludes the paper with possible future work.
%%%%%%%%%%%%%%%%%%%%%%%%%%%%%%%%%%%%%%%%%%%%%%%%%%%%%%%%%%%%%%%%%%%%%%%%%%%%%%%%%%%%%%%%%%%%%%%%%%%%%%%%

\section{\textbf{Related Work}}
\label{related_work}
%%%%%%%%%%%%%%%%%%%%%%%%%%%%%%%%%%%%%%%%%%%%%%%%%%%%%%%%%%%%%%%%%%%%%%%%%%%%%%%%%%%%%%%%%%%%%%%%%%%%%%%%
In this section, we present a summary of an extensive literature review on related work done in SLAM applications, using different kinds of techniques. Throughout the years, many techniques and algorithms have been introduced to address the SLAM problem \cite{durrant2006simultaneous,bailey2006simultaneous}. Popular techniques include visual-SLAM, magnetic-SLAM, and Wifi-SLAM. 
%%%%
Visual-SLAM methods, utilize RGB-D cameras like Kinect and Tango \cite{engelhard2011real} to acquire 3D models of the environment.
Magentic-SLAM systems, exploit magnetic field for localization and mapping of mobile robots \cite{kim2007slam}. 
These techniques use a Gaussian process to model the magnetic field intensity 
and a particle filtering to estimate the pose of mobile robot \cite{vallivaara2011magnetic}.

Wifi-SLAM \cite{ferris2007wifi, Huang_wifi_slam_11, jirkuu2016wifi} techniques use the radio signal and motion data of the device for localization and signal strength mapping in unknown environment. 
Authors in \cite{signal_slam} extended this approach to include more sensory information, for example Bluetooth, LTE, and magnetic fields. 
The traditional fingerprinting-based approaches need a surveying phase 
to collect the radio measurements in an environment and annotate them 
with locations which are measured by an external reference system. 
This process is highly labour and time intensive, 
thus limiting the applicability of fingerprinting-based approaches. 
To overcome such limitations, researchers are moving towards finding low cost methods to generate the radio map. 
Such methods involve little human intervention \cite{yang2012locating, subbu2014analysis} as they leverage the in-built sensing capabilities of mobile phones. 
Hence, with SLAM technique, the hassle of site surveying can be avoided, and radio map can be updated conveniently whenever needed. 

When the indoor environment becomes huge, generating the radio map with single mobile device becomes time consuming. 
The power of crowd-sensing comes into play in this scenario. 
Mobile crowd-sensing is a popular computing paradigm which enables ubiquitous mobile devices to collect sensing data at large scales \cite{dong2017unleashing, marakkalage2018understanding}. 
Crowd-sensing techniques can be utilized to unleash the potential of mobile phones of people who move inside the indoor environment \cite{faragher2012opportunistic}. 
Moreover, such systems do not require prior knowledge of floor plans or locations of wireless transmitters.
Prior researches have harnessed the power of crowd-sensing to reconstruct indoor floorplans by combining user mobility traces, visual \cite{gao2014jigsaw}, and Wifi fingerprinting for indoor Wifi monitoring (Pazl) \cite{radu2013pazl}. 
In our system, we combine crowdsensed RSS from Wifi APs and Google Tango trajectory information. 
The system implementation is explained in detail in section \ref{system_overview}

\begin{figure}
\centering
\includegraphics[width=0.7\textwidth]{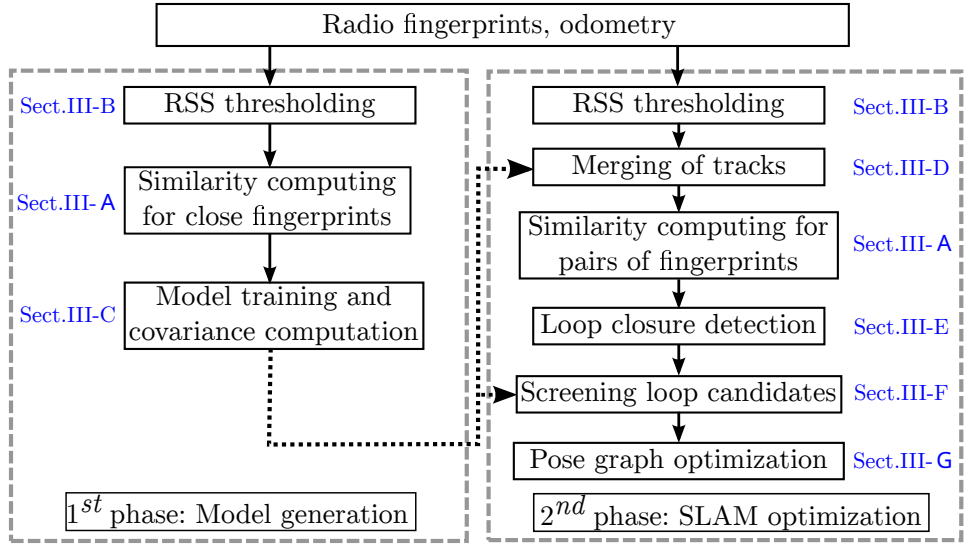}
\caption{
System overview. }
\label{fig:system_overview}
\end{figure}

%%%%%%%%%%%%%%%%%%%%%%%%%%%%%%%%%%%%%%%%%%%%%%%%%%%%%%%%%%%%%%%%%%%%%%%%%%%%%%%%%%%%%%%%%%%%%%%%%%%%%%%%
\section{\textbf{Crowd-sensing SLAM based on Pose Graph Optimization}}
\label{system_overview}	
%%%%%%%%%%%%%%%%%%%%%%%%%%%%%%%%%%%%%%%%%%%%%%%%%%%%%%%%%%%%%%%%%%%%%%%%%%%%%%%%%%%%%%%%%%%%%%%%%%%%%%%%
%The problem we want to solve is actually SLAM.
We present a novel approach that incorporates radio fingerprint measurement and motion information for crowd-sensing SLAM in an unknown infrastructure. 
Particularly, our approach does not need any prior information about location of the access point 
nor does it require a labor-intensive phase to collect the measurements in advance.
Our approach features a cost-effective alternative for trajectory estimation of multiple users in unknown environments. 
The estimated trajectory can be used to create a radio map of the environment for the localization of a user afterwards.

The goal is to infer the entire trajectories from observations taken at different times without any prior knowledge of the environment. 
The problem addressed here is known as SLAM, 
with a variety of solutions have been proposed in the literature \cite{RFM-SLAM, Thrun_Probabilistic_robotics, slam_trends_2016, ceres-solver}.
In Graph-based SLAM \cite{Thrun_Probabilistic_robotics}, a graph is constructed from the raw sensor measurements.
Nodes in the graph are represented by the pose of the user.
The edge between two nodes encodes the spatial constraint that links them.
%A constraint consists a probability  distribution  over  the  relative  transformations between the two poses. 
%These transformations are either odometry  measurements  between  sequential  robot  positions  
%or  are determined  by  aligning  the  observations  acquired  at  the  two robot  locations.  
A constraint is either sequential odometry measurement or 
the transformation (i.e., loop closure) determined by aligning the measurements at two non-consecutive poses. 
Since the observations are noisy, 
all constraints are additionally parameterized with a certain degree of uncertainty. 
The problem turns into finding the best configuration of the poses to minimize these constraints. 

Loop closure is essential for a SLAM system, 
since it allows to correct the accumulated odometric errors and create a consistent map of the environment.
It represents a situation that users have revisited a previously observed location.
The detection of a loop can be 
achieved by feature or scan matching algorithms using visual cameras or laser range finders \cite{Taketomi2017, SujiwoATNE16}. 

Formally, we denote $\x=\{ \x_{1},...,\x_{T} \}^\intercal$ as the path of the user to be estimated up to time $T$,
where $\x_t= (x_t,y_t,\theta_t)$ is the global 2D location and heading of the user at time $t$. 
Let $\z_{ij}$ and $\Sigma_{ij}$ denote the mean and covariance of a measurement (i.e., constraint) between node $i$ and $j$.
Let $\C$ be the set containing indices of all pairs of constraints in the graph.
$\hat{\z}_{ij}(\x_i,\x_j)$ is the prediction of a measurement based on the current configuration of node $i$ and $j$.
The graph-based SLAM aims to find the best configuration of $\x$ to minimize the following equation:
\begin{equation}
\sum_{(i,j) \in \C} {(\z_{ij}-\hat{\z}_{ij}(\x_i,\x_j))}^\intercal  {\Sigma}_{ij}^{-1} (\z_{ij}-\hat{\z}_{ij}(\x_i,\x_j))
\end{equation}
In particular for graph-based SLAM, 
$\z_{ij}$ is expressed as a rigid-body transformation between node $i$ and $j$. 
%Each RSS values is associated with the mac address of the access point.
Given a signal strength measurement from an AP, 
it is straightforward to know if an area has been visited by a user, 
since each reported RSS value is associated with a unique MAC address. 
However, estimating the precise transformation between two observations turns out to be tricky, 
since RF signal neither reports distance nor bearing, 
and the detection range of an AP can be up to 50 meters, 
which is usually much larger than accumulated error of a pedestrian dead reckoning system.
This is quite different from laser range finder, 
where the transformation can be estimated by matching of two laser scans \cite{Lu_millos_1997}.

A number of researchers use analytical models to predict propagation of radio signal over a distance.
Many factors (examples include multiple path or obstruction from obstacles) will distort the propagation of the signal, 
and it is not practical to model all these aspects.
Instead of modeling them explicitly, this paper represents the location with radio fingerprints.
This is motivated by the fact that the similarity of two fingerprints highly depends on the locations of the two measurements.
%Exploiting the similarity prevents us from rss modeling and measuring the location of the access points in advance.

We claim a loop closure if the similarity between two radio measurements at times $i$ and $j$ reaches a threshold $\vartheta_s$. 
We then infer that their positions are the same
and add a constraint $\z_{ij}=(0, . . . ,0)$ to the graph. 
Actually, 
the two locations are unlikely to be exactly the same, which will introduce a small amount of error to the loop closure.
The error can be compensated by associating a covariance $\Sigma_{ij}$ to the constraint. 
A choice of this can be a diagonal matrix with small values on the main diagonal.
Our solution is a careful examination of the uncertainty of a loop closure based on the degree of similarity in a training phase. 
Based on the training data, we automatically learn a nonparametric model to feature the variance of the distance given the similarity of two location fingerprints.
The following subsections will describe each component of our proposed solutions in details.
%to address the issue of how to recognize a location where the user has previously visited. 

%By comparing the degree of similarity between two fingerprints at distant timestamps, 
%we can infer that a location is being visited again. 
%As a result, a constraint edge will be added to the pose graph.
%Additionally, based on training data, 
%we learn a nonparametric model to feature the degree of similarity conditioned on the distance of two locations.
%The process does need to know the placement of the access points 
%and is acquired on fly and is adaptive to different environment.
%The next section will address the issue of how to recognize a location that the user has been earlier.

%%%%%%%%%%%%%%%%%%%%%%%%%%%%%%%%%%%%%%%%%%%%%%%%%%%%%%%%%%%%%%%%%%%%%%%%%%%%%%%%%%%%%%%%%%%%%%%%%%%%%%%%
\subsection{Radio Fingerprints and the Similarity}
\label{sect_rf_fingerprint}
%%%%%%%%%%%%%%%%%%%%%%%%%%%%%%%%%%%%%%%%%%%%%%%%%%%%%%%%%%%%%%%%%%%%%%%%%%%%%%%%%%%%%%%%%%%%%%%%%%%%%%%%
RF fingerprinting represents a location with a set of radio signals from 
the nearby transmitters, 
for example Wifi APs, RFID, and Bluetooth. 
These fingerprints are considered to be robust against location-specific distortions 
as compared to the propagation model-based approaches,
since they can capture the unpredicted characteristics of the existing radio infrastructure. 
This is quite similar to appearance-based approach, where the scene is represented by a number of visual features \cite{gao2018ldso}.
Extracting this kind of features involves a large amount of computation, 
while this process can be ignored for the radio fingerprint
since each AP is unique and can be used as a feature to identify a place. 

We represent a fingerprint at time $t$ as a pair $\F_t={(\f_t,\x_t)}$.
$\f_t$ consists of the signal strength from $L$ access points measured at location $\x_t$: $\f_t=\{ f_{t,1},...,f_{t,L} \}$.
The similarity function $\SIM(\F_i,\F_{j})$ returns a positive scalar value, representing the similarity between two vectors $\F_{i}$ and $\F_{j}$.
In our experiments, 
we adopt the cosine similarity which has been extensively used by many researchers \cite{RanArtur_IROS_2012, vorst2010isr}:
\begin{equation}
\s_{ij}=\SIM(\F_{i},\F_{j})=
\frac{\sum_{l=1}^L{f_{i,l} f_{j,l}}}{\sqrt{\sum_{l=1}^L{\left(f_{i,l}\right)^{2}}}\sqrt{\sum_{l=1}^L{\left(f_{j,l}\right)^{2}}}}
\end{equation}
We refer the readers to \cite{he2016wi, vorst2011rfidta} for a comparison of different similarity measures.

%%%%%%%%%%%%%%%%%%%%%%%%%%%%%%%%%%%%%%%%%%%%%%%%%%%%%%%%%%%%%%%%%%%%%%%%%%%%%%%%%%%%%%%%%%%%%%%%%%%%%%%%
\subsection{RSS Thresholding}
\label{sect_rss_threshold}
%%%%%%%%%%%%%%%%%%%%%%%%%%%%%%%%%%%%%%%%%%%%%%%%%%%%%%%%%%%%%%%%%%%%%%%%%%%%%%%%%%%%%%%%%%%%%%%%%%%%%%%%
The time required to compute the similarity 
increases linearly with the number of detected APs in the two fingerprints. 
The computational cost can be very high in densely AP covered scenarios such as indoor stadium or airport.
A large amount of computational time can be saved if the size of the measurements can be reduced.
Therefore, we propose to filter out the observations whose RSS values are below a threshold $\vartheta_r$.

Thresholding prunes observations with smaller RSS values, which represent spurious readings. 
In addition, larger RSS values indicate a location close to the access point with more confidence. 
They are expected to better constrain the location of the user.
In the experimental section, 
we show that thresholding technique provides better accuracy, 
but with less computational time.

\begin{figure}
  \centering
    \subfigure[Experimental snapshot]{
\label{fig:environment}
    \includegraphics[height=0.4\textwidth]{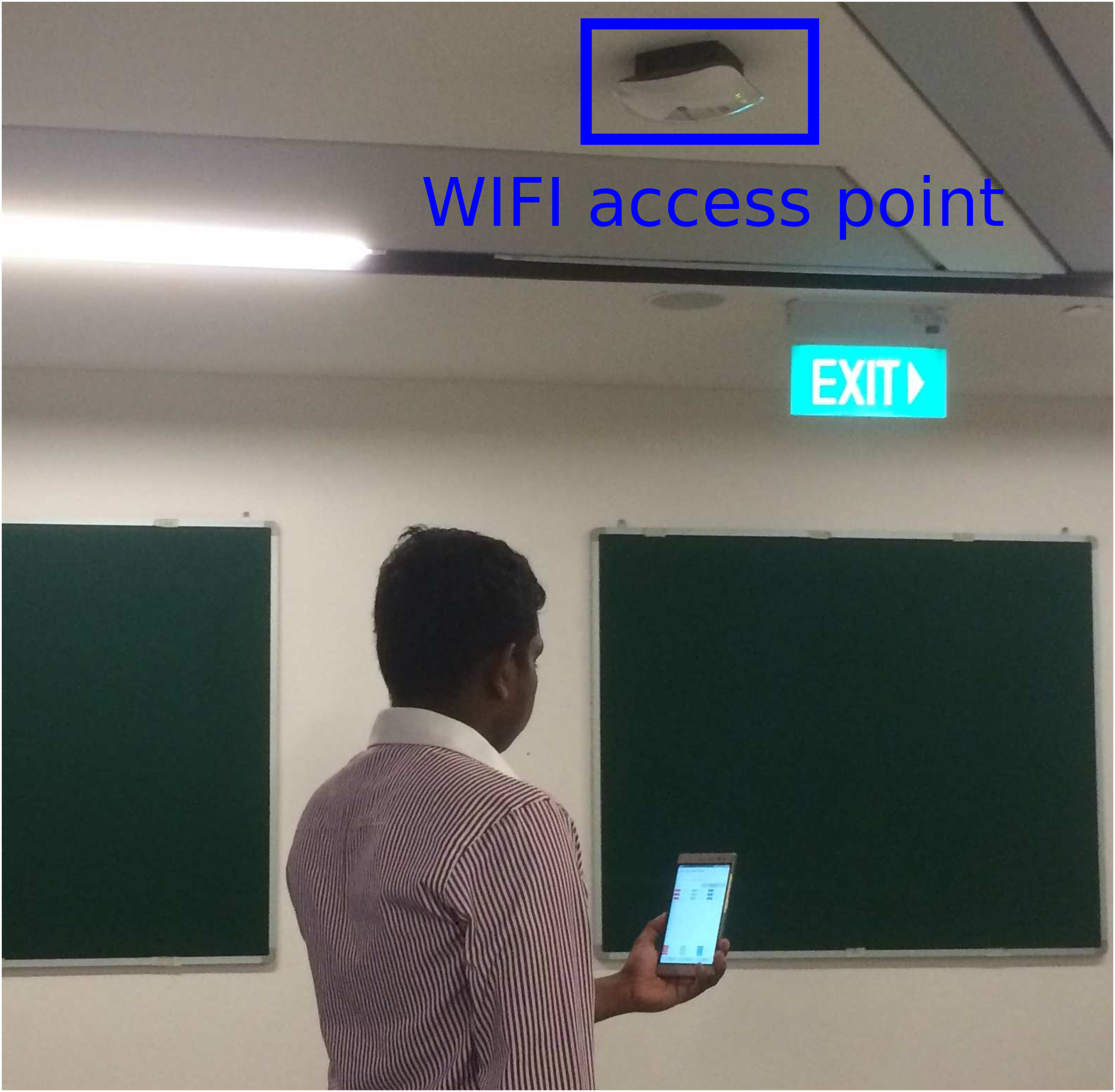}
    }   
    \hspace{-0.4cm}
  \subfigure[Similarity and the distance variance]{
\label{fig:sim_distance}
        \includegraphics[height=0.4\textwidth]{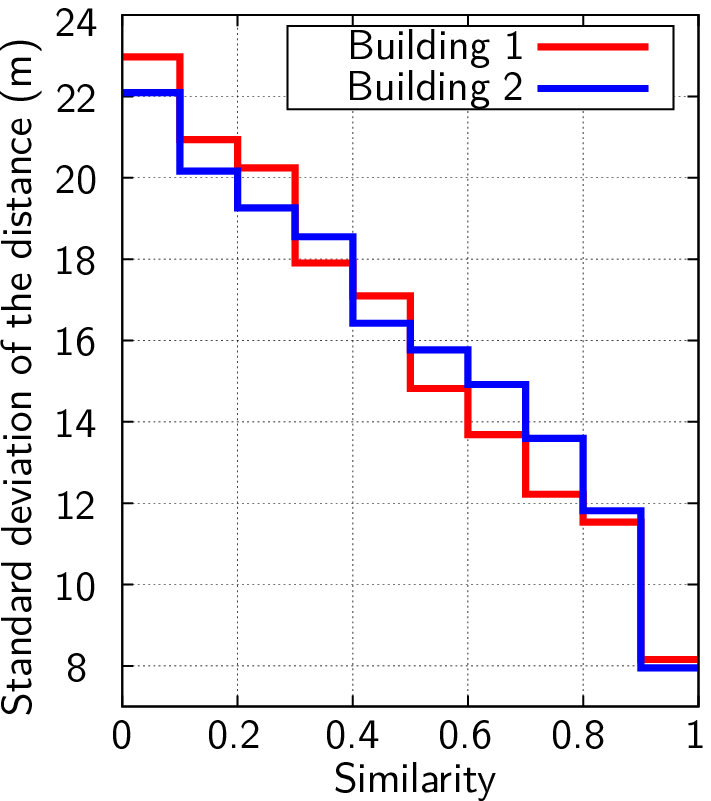}
        }
   \caption[Similarity]
{(a) One experimental snapshot;
(b) Similarity and the distance variance in two different buildings.}
\label{fig:sim_distance}
\end{figure}

%%%%%%%%%%%%%%%%%%%%%%%%%%%%%%%%%%%%%%%%%%%%%%%%%%%%%%%%%%%%%%%%%%%%%%%%%%%%%%%%%%%%%%%%%%%%%%%%%%%%%%%%
\subsection{Model Training}
\label{sect_model_training}
%%%%%%%%%%%%%%%%%%%%%%%%%%%%%%%%%%%%%%%%%%%%%%%%%%%%%%%%%%%%%%%%%%%%%%%%%%%%%%%%%%%%%%%%%%%%%%%%%%%%%%%%
An uncertainty estimation of the constraint is required for all edges in a SLAM graph. 
For odometry-based edges, this parameter is obtained from the motion model. 
We now need to derive a model to feature the uncertainty of the observation-based edges.
Our solution is to train such a model by passing over the sensor data which is recorded at hand as shown in Figure \ref{fig:system_overview}.

Although the error from odometry accumulates in the long term, 
it is sufficiently small for a short time of duration. 
In this work, we assume odometry is accurate enough 
if the distance traveled is less than 30 m. 
This value should be modified based on the accuracy of dead reckoning system, 
but according to our knowledge 30 m is a suitable value for most inertial tracking platforms.
Therefore, we compute the degree of similarity for fingerprint pairs whose position is close. 
These values are annotated with the distance of the two locations. 
As a result, we will get a set of $K$ training samples: $\{\s_k,d_k\}_{k=1}^{K}$, 
where $\s_k$ is the similarity and $d_k$ is the distance of the fingerprint pair. 
We then train a model which features the variance of distance given a similarity by binning. 
%The uncertainty of a similarity $var(s)$ is the approximated 
%by the variance of the samples sites in the interval $b$ .
%From the training samples, 
%we want get the variance of distance $d$ at a specific similarity values $s$, when between two nodes $i$ and $j$.
That is, for a similarity value $\s$, 
we compute the variance of the samples that sites in the small interval $b$ around $\s$:
\begin{equation}
var{(\s)}=\frac{1}{ c (b (\s)) } \sum_{k \in b(\s)} { d_k}^2
\end{equation}
where $c(b)$ counts the number of samples in interval $b$. 
Although binning is a simple way for smoothing, 
the computation is efficient, 
since assigning a sample into a bin is straightforward. 
The resulted variance is stored in a look up table 
which could be used in the second stage of SLAM, as shown in Figure \ref{fig:system_overview}. 

%%%%%%%%%%%%%%%%%%%%%%%%%%%%%%%%%%%%%%%%%%%%%%%%%%%%%%%%%%%%%%%%%%%%%%%%%%%%%%%%%%%%%%%%%%%%%%%%%%%%%%%%
\subsection{Merging Two Tracks at Different Times}
\label{track_merge}
%%%%%%%%%%%%%%%%%%%%%%%%%%%%%%%%%%%%%%%%%%%%%%%%%%%%%%%%%%%%%%%%%%%%%%%%%%%%%%%%%%%%%%%%%%%%%%%%%%%%%%%%
To incorporate the crowd-sensed measurements from multiple users, 
it is necessary to merge tracks from different users. 
This allows to perform loop closure detection among different tracks and make a full use of the crowd-sensed data. 
%See section C we only infer a loop closure when the distance or orientation change over a threshold. 
%In the graph-based SLAM,
%we will have to find an edge that connects different tracks. 
We assume the orientation of the device is known or can be approximately estimated by an external system. 
This is quite reasonable for the following reasons:
compass can directly offer the orientation if the placement of the device is fixed.
For arbitrary placements, 
authors in \cite{pdr_arbitrary_placement} proposed an approach to determine the walking direction of a pedestrian 
by projecting the displacement vectors onto a plane.
%An alternative way would be to formulate this as a state estimation problem, 
%which can be solved efficiently by state-of-the-art probability-based approaches 
%such as the Kalman filter or the particle filter \cite{Thrun_Probabilistic_robotics} \cite{Yassin_ieee_tutorials_2016}. 

In our experiment, 
we assume every user started from the same location. 
Thus, an edge is added to link the first node between two tracks. 
The transformation along $x$ and $y$ is set to be zero 
and the orientation $\theta$ is approximated by the orientation difference between the two nodes. 
The covariance along $x$ and $y$ is determined by the covariance look up table as detailed in Sect.\ref{sect_model_training} 
and orientation covariance is set to a large value (i.e., 1000), 
which means that we do not have any knowledge about the orientation of the two poses based on only RF observations. 
In case that the relationship between the initial locations is unknown, 
it can be determined by state-of-the-art probability-based approaches 
such as the Kalman filter or the particle filter \cite{Thrun_Probabilistic_robotics, Yassin_ieee_tutorials_2016}. 
%In case that the initial location of a user is unknown, 
%the pose estimation can be 
%efficiently solved by state-of-the-art probability-based approaches 
%such as the Kalman filter or the particle filter \cite{Thrun_Probabilistic_robotics} \cite{Yassin_ieee_tutorials_2016}. 
%%%%%%%%%%%%%%%%%%%%%%%%%%%%%%%%%%%%%%%%%%%%%%%%%%%%%%%%%%%%%%%%%%%%%%%%%%%%%%%%%%%%%%%%%%%%%%%%%%%%%%%%
\subsection{Finding Loop Closure Candidates}
\label{sect_finding_loop_candidate}
\begin{figure}
\centering
\includegraphics[width=0.7\textwidth]{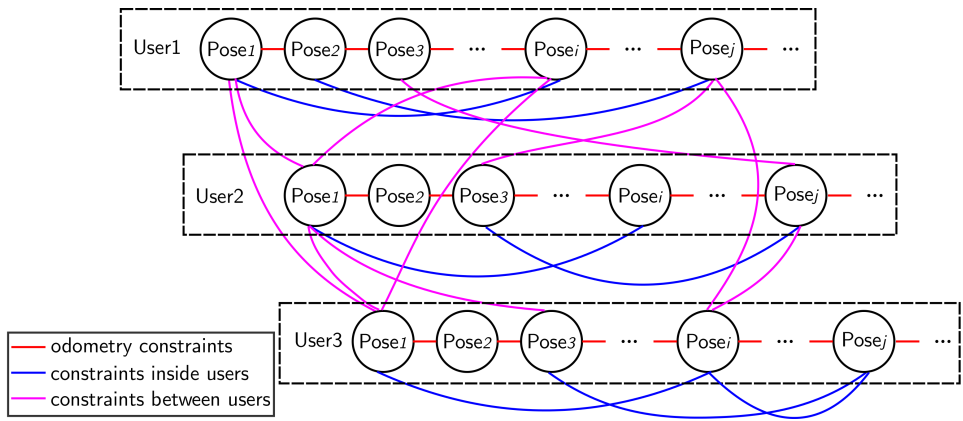}
\caption{Example of constraints inferred from the measurements taken by three users. 
}
\label{fig:illustrate_constraints}
\end{figure}

%%%%%%%%%%%%%%%%%%%%%%%%%%%%%%%%%%%%%%%%%%%%%%%%%%%%%%%%%%%%%%%%%%%%%%%%%%%%%%%%%%%%%%%%%%%%%%%%%%%%%%%%
We first compute the relative distance and orientation of two fingerprints $\F_i$ and $\F_j$. 
If they are smaller than pre-defined thresholds, 
we compute the similarity $\s_{ij}$ between $\F_i$ and $\F_j$.
We add a tuple $<i,j,\s_{ij}>$ as a candidate of loop closure if the similarity exceeds a threshold $\vartheta_s$, 
which is one of the few parameters that has to be supplied in our approach. 
The impact of $\vartheta_s$ on the performance is not too critical, as shown in our experiments. 
In most cases, $\vartheta_s=0.8$ gives a good accuracy. 
We reject the similarity with values smaller than $\vartheta_s$, to avoid false positive loop closures.
It is worth mentioning that this process is applied not only to the nodes in the same track, 
but also to the nodes between two tracks to leverage crowd sourced measurements collected by multiple users. 
Constraints inferred from odometry and radio measurements from three users 
are illustrated in Figure \ref{fig:illustrate_constraints}.
%We caculate the similarity bet
%we could obtain the $k$ reference fingerprints ${\f_{\pi(1)},...,\f_{\pi(k)}}$
%whose similarities best match the measured RSS sequence $\g_t$. These fingerprinting is used to infer further loops.
%%%%%%%%%%%%%%%%%%%%%%%%%%%%%%%%%%%%%%%%%%%%%%%%%%%%%%%%%%%%%%%%%%%%%%%%%%%%%%%%%%%%%%%%%%%%%%%%%%%%%%%%
\subsection{Screening Loop Closure Candidates}
\label{sect_screening}
%%%%%%%%%%%%%%%%%%%%%%%%%%%%%%%%%%%%%%%%%%%%%%%%%%%%%%%%%%%%%%%%%%%%%%%%%%%%%%%%%%%%%%%%%%%%%%%%%%%%%%%%
Incorrect loop closure is catastrophic to a SLAM system, 
as it will ruin the consistency of a trajectory and the map. 
We perform the following check of a loop candidate 
before we finally add it to the graph for optimization:
\begin{itemize}
\item We discard a loop $<{i,j,\s_{ij}}>$ if the difference of $i$ and $j$ is smaller than $M$ 
to prevent from detecting loops when the user is still at the same location;
\item For each constraint $<{i,j,\s_{ij}}>$, 
we check if there exists another candidate $<{i,k,\s_{ik}}>$ that lies within $\vartheta_w$ meters before or after $j$ in the track.
If yes, the one with low similarity value will be removed.
\end{itemize}

%The final loop closure should go through some geometric consistency check. 
One could think about other heuristic approaches to filter out the suspicious loop closures. 
For example, authors in \cite{GalvezIROS11} propose to group the loop closures 
that close in time and check the temporal consistency with previous scenes for robust loop closure detection.
However, this goes beyond the focus of this paper, 
hence, we evaluate the performance of the system without further loop consistency check.

After screening, 
the remaining constraints will be added as edges to the graph. 
We set the transformation of the edge to zero for both translation and orientation. 
Covariance of the translation can be found in the look up table we computed in Section \ref{sect_model_training}. 
The orientation covariance is set to 1000. 

%%%%%%%%%%%%%%%%%%%%%%%%%%%%%%%%%%%%%%%%%%%%%%%%%%%%%%%%%%%%%%%%%%%%%%%%%%%%%%%%%%%%%%%%%%%%%%%%%%%%%%%%
\subsection{Pose Graph Optimization}
\label{pose_graph_optimization}
%%%%%%%%%%%%%%%%%%%%%%%%%%%%%%%%%%%%%%%%%%%%%%%%%%%%%%%%%%%%%%%%%%%%%%%%%%%%%%%%%%%%%%%%%%%%%%%%%%%%%%%%
We finally optimize the graph consisting of poses and constraints based on the pose graph optimization technique. 
%We choose g2o to achive the optimization, which is an open-source framework, 
We choose Levenberg-Marquardt in g2o as the implementation \cite{kuemmerle11icra}. 
The algorithm is freely available 
and is proven to be one of the state-of-the-art SLAM algorithms. 

\begin{figure}
\centering
\includegraphics[width=0.6\textwidth]{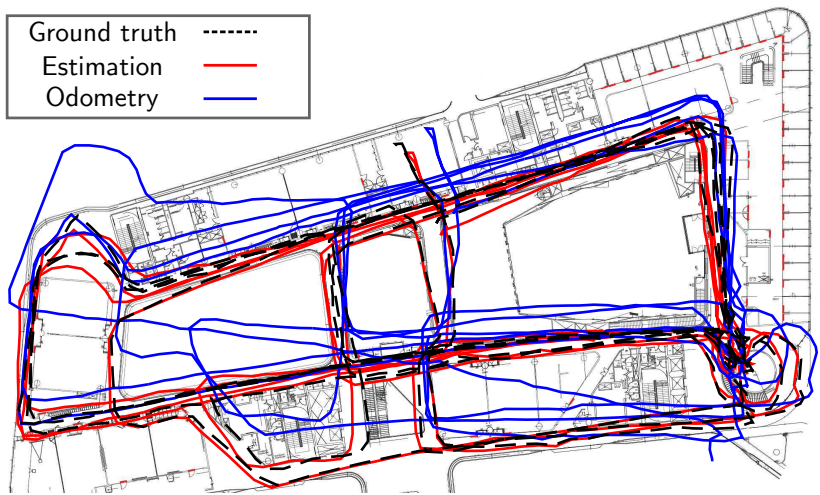}
\caption{Estimated path, ground truth, and the raw odometry provided by Tango. 
}
\label{fig:floor_plan}
\end{figure}

\begin{figure*}
  \centering 
       \subfigure[Track1]{
\label{fig:trajectory}
        \includegraphics[height=0.4\textwidth]{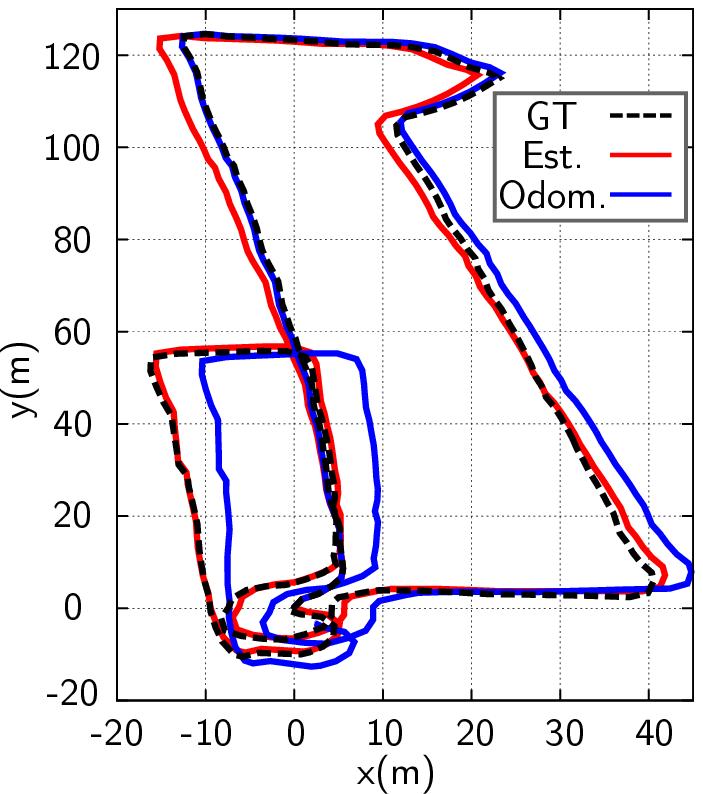}
        }
        \hspace{-0.3cm}
     \subfigure[Track2]{
\label{fig:trajectory}
        \includegraphics[height=0.4\textwidth]{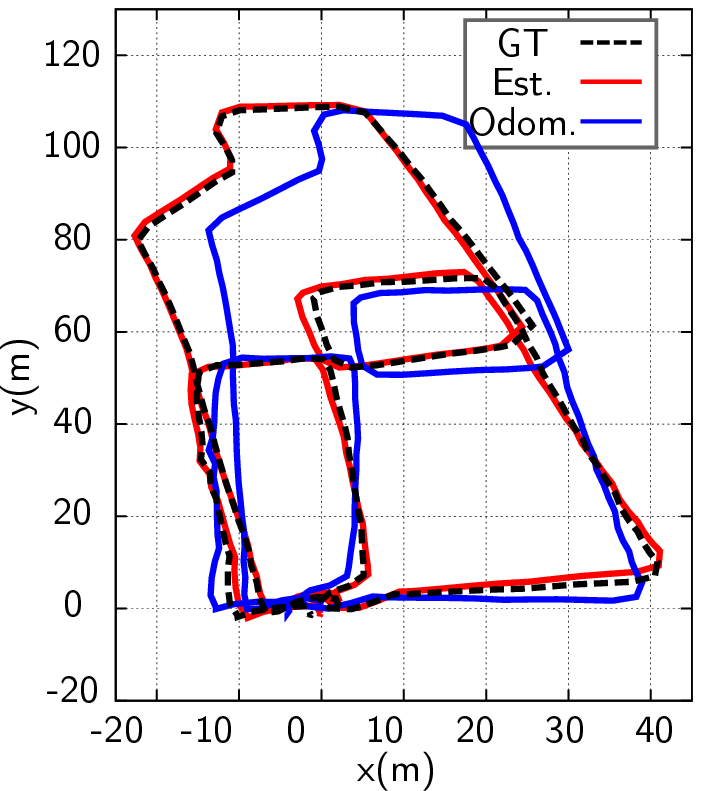}
        }
        \hspace{-0.3cm}
 \subfigure[Track3]{
\label{fig:tracking_error}
        \includegraphics[height=0.4\textwidth]{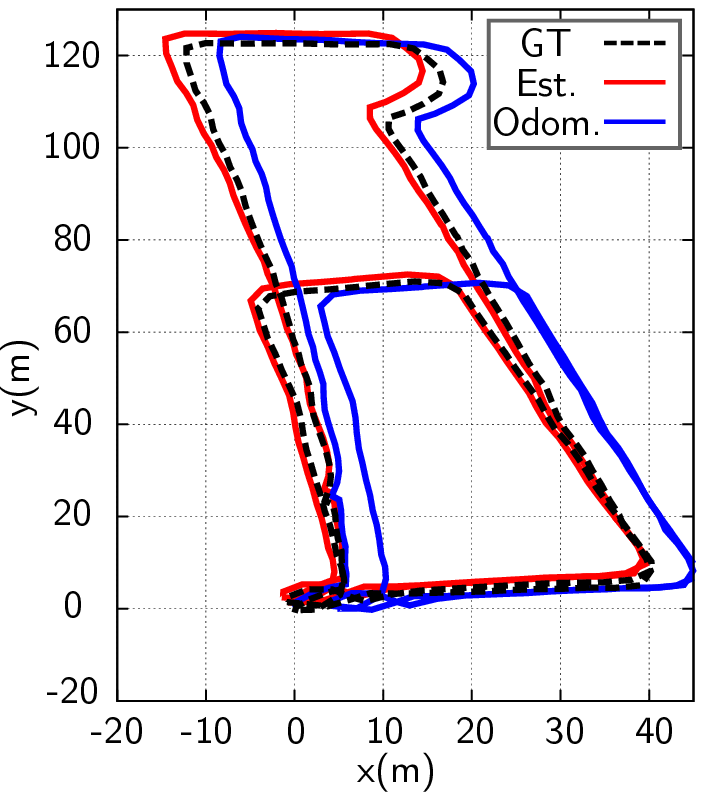}
        }
        \hspace{-0.3cm}
        \subfigure[Track4]{
\label{fig:impact_of_different_imu_noise}
    \includegraphics[height=0.4\textwidth]{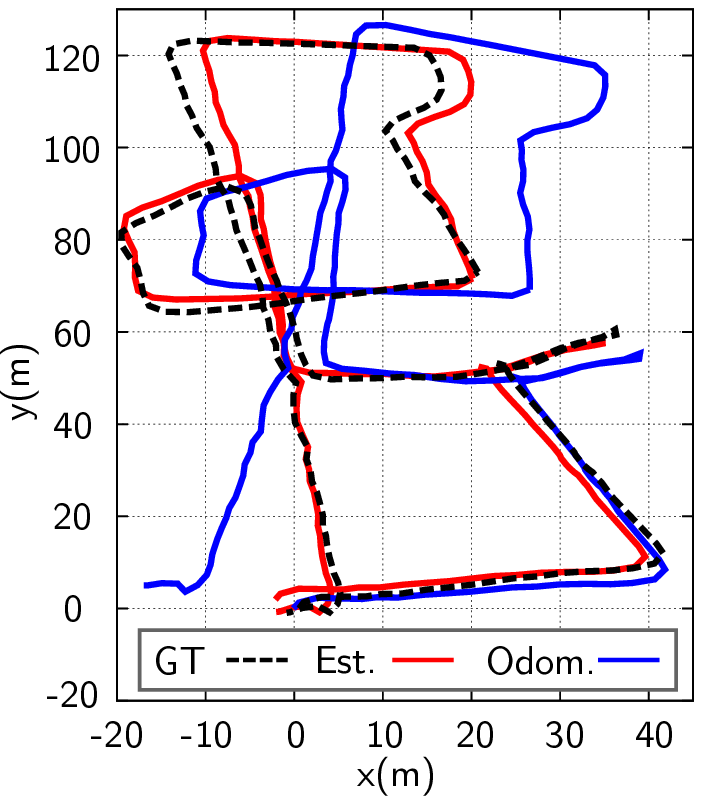}
    }    
   \caption[Tracking of fusion.]
{Estimation, ground truth from point cloud-based SLAM, and the raw odometry of individual tracks.
}
\label{fig:example_track}
\end{figure*}
%%%%%%%%%%%%%%%%%%%%%%%%%%%%%%%%%%%%%%%%%%%%%%%%%%%%%%%%%%%%%%%%%%%%%%%%%%%%%%%%%%%%%%%%%%%%%%%%%%%%%%%%
\section{\textbf{Experimental Results}}
\label{experimental_evaluations}
%%%%%%%%%%%%%%%%%%%%%%%%%%%%%%%%%%%%%%%%%%%%%%%%%%%%%%%%%%%%%%%%%%%%%%%%%%%%%%%%%%%%%%%%%%%%%%%%%%%%%%%%
%%%%%%%%%%%%%%%%%%%%%%%%%%%%%%%%%%%%%%%%%%%%%%%%%%%%%%%%%%%%%%%%%%%%%%%%%%%%%%%%%%%%%%%%%%%%%%%%%%%%%%%%
\subsection{Experimental Details}
\label{Implementation Detail}
%%%%%%%%%%%%%%%%%%%%%%%%%%%%%%%%%%%%%%%%%%%%%%%%%%%%%%%%%%%%%%%%%%%%%%%%%%%%%%%%%%%%%%%%%%%%%%%%%%%%%%%%

We program the Tango phone (with Android 6.0.1) 
to receive the signal strength from APs and simultaneously upload them to the server every five seconds.  
Meanwhile, the pose as well as the point cloud from the device are sent to the server every one second. 
We evaluated the performance of our approach on the third floor of our campus building with a size of 130 m$ \times $70 m, as shown in Fig. \ref{fig:floor_plan}.
This environment consists of concrete walls, corridors, soft room partitions, and wide open space. 
A person was asked to hold the phone and walk around the environment along different paths with a normal walking speed. 
In total, we recorded four tracks at different times. 
For each track, we asked the user to start from the same position. 
The total distance traveled is 1533 meters with a duration of 3740 seconds and a number of 
2211 unique MAC addresses are detected.
This results in four log files consisting of 748 measurements, with 207, 170, 181, and 190 in each track respectively. 
A snapshot of the environment is shown in Fig. \ref{fig:environment}.

%%%%%%%%%%%%%%%%%%%%%%%%%%%%%%%%%%%%%%%%%%%%%%%%%%%%%%%%%%%%%%%%%%%%%%%%%%%%%%%%%%%%%%%%%%%%%%%%%%%%%%%%
\subsection{Point Cloud-based SLAM as Ground Truth}
\label{ground_truth}
%%%%%%%%%%%%%%%%%%%%%%%%%%%%%%%%%%%%%%%%%%%%%%%%%%%%%%%%%%%%%%%%%%%%%%%%%%%%%%%%%%%%%%%%%%%%%%%%%%%%%%%%
We compared our results against a point cloud-based GraphSLAM. 
We implemented loop closure detection based on point cloud using the open source point cloud library (PCL) \cite{pcl_icra_2011}.
We identify the Harris keypoints in a pair of point clouds and compute the corresponding SHOT descriptors \cite{Alexandre123ddescriptors}. 
We match these descriptors with KNN and find an initial transformation using SVD (singular value decomposition).
The transformation is further refined by ICP (iterative closest point).
If the number of matched points exceeds a threshold (half size of the points),
a loop closure is confirmed and added to the graph as constraints. 
%We further compute the covariance of the transformation following Lu and Millos \cite{Lu_millos_1997}.
%And the transformation as well as the covariance (followed by Lu and Millos) 
We treat this as the ground truth to evaluate the accuracy of our approach. 

The accuracy is shown by the root mean square error (RMSE) of the distance between the ground truth and our estimation.
Our experiments show that we are able to achieve an accuracy of 1.74 meters over an area of approx. 9000 square meters, as shown in Figure \ref{fig:floor_plan}.
The optimized track is annotated with the RF measurement 
and can serve as the radio map for the positioning of other users.

\begin{figure}
\centering
\includegraphics[width=0.7\textwidth]{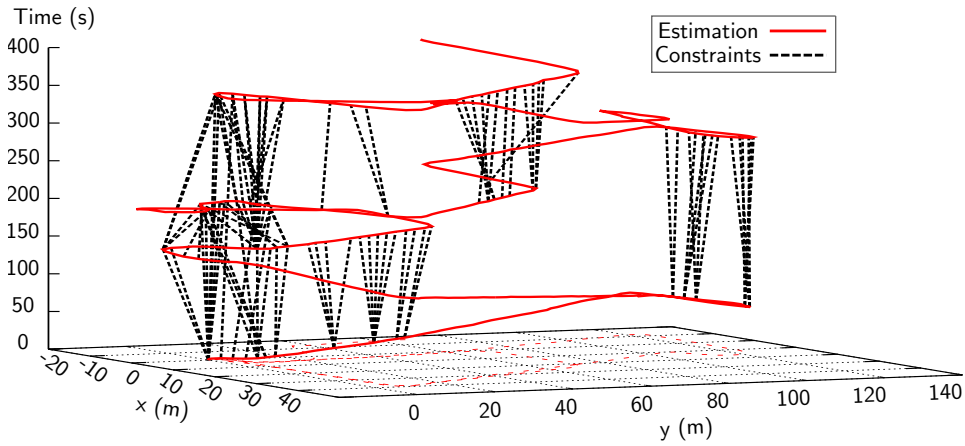}
\caption{
Part of the estimated track over time 
and constraints inferred with a similarity threshold of $\vartheta_s=0.8$. }
\label{fig:constraints}
\end{figure}

\begin{table}[]
\centering
\caption{accuracy (mean, standard deviation, median, and maximum) in meters, average number of mac addresses detected, and the computational time of loop closure detection under the impact of different RSS threshold $\vartheta_r$.}
\label{table_rss}
\begin{tabular}{|c|c|c|c|c|}
\hline
$\vartheta_r$ & \begin{tabular}[c]{@{}c@{}}Mean$\pm$ \\ Std. Dev.\end{tabular} & Maximum & \begin{tabular}[c]{@{}c@{}}Number \\ of MAC\end{tabular} & \begin{tabular}[c]{@{}c@{}}Comp. \\ time (s)\end{tabular}   \\ \hline
-90       & 2.47$\pm$2.47   &   11.84  &     163.2 &     36.53                 \\ \hline
-80       & 1.94$\pm$1.41     & 7.34         &   106.8             &      21.9      \\ \hline
-70       &    1.74$\pm$0.89        &       4.05   & 39.4               &     7.35 \\ \hline
-60       &  2.29$\pm$1.72         &  10.06      &    14.2                  &     2.61 \\ \hline
-50       &       3.86$\pm$2.33    &  12.56        &     2.8                &     0.52 \\ \hline
-45       &  3.50$\pm$3.79        &  19.28        &        0.8              &    0.21 \\ \hline
 \begin{tabular}[c]{@{}c@{}}Odom.\\  only \end{tabular}     &  4.99$\pm$4.26   &  22.13        &    N/A                  &  N/A   \\ \hline

\end{tabular}
\end{table}

%%%%%%%%%%%%%%%%%%%%%%%%%%%%%%%%%%%%%%%%%%%%%%%%%%%%%%%%%%%%%%%%%%%%%%%%%%%%%%%%%%%%%%%%%%%%%%%%%%%%%%%%
\subsection{Impact of Different RSS Threshold $\vartheta_r$}
\label{evaluation_rss_threshold}
%%%%%%%%%%%%%%%%%%%%%%%%%%%%%%%%%%%%%%%%%%%%%%%%%%%%%%%%%%%%%%%%%%%%%%%%%%%%%%%%%%%%%%%%%%%%%%%%%%%%%%%%
In the first series of experiments, we examined the accuracy under the impact of RSS thresholding technique. 
The results are listed in Table \ref{table_rss}. 
We set the similarity threshold $\vartheta_s=0.8$ and use a binning size of $b=0.1$. 
According to the distribution of the RSS, 
we vary $\vartheta_r$ from -90 to -45 to analyze its impact on the positioning accuracy.
As can be seen from Table \ref{table_rss}, 
we maintain a good accuracy with a small threshold (i.e., $\vartheta_r=-70$ or $-60$), 
while the computational time decreases considerably. 
%In densely AP covered environment, 
%the access points are randomly distributed in an environment.
%A large distance or the occlusions from the environment can both lead to a low signal strength. 
For example, a threshold of -70 will reduce the computational time from 36.53 seconds to 7.35 seconds as compared to the threshold of -90. 
At the same time, the accuracy even increases by 0.73 meters.
The reason for this can be explained as: 
A large number of occlusions are expected in an indoor environment, 
resulting in a huge variance of the received signal strength. 
In densely AP covered areas, 
many APs report very low signal strength due to the regular occlusions in indoor environments, 
which will reduce the overall similarity between two fingerprints as well as the final accuracy we achieved. 
A suitable threshold will give a good accuracy, as it filters out the suspicious signals. 
However, a threshold larger than -50 leads to a bad result (for example, 3.86 meters of error at running time of 0.52 seconds with $\vartheta_r=-50$). 
The estimation, ground truth, and odometry of individual tracks are visualized in Figure \ref{fig:example_track}.
A part of estimated trajectory and the constraints inferred are shown in Figure \ref{fig:constraints}.

\begin{table}[]
\centering
\caption{accuracy (mean and standard deviation, median, and maximum) in meters, and the number of constraints inferred under the impact of different similarity threshold $\vartheta_s$.}
\label{table_sim}
\begin{tabular}{|c|c|c|c|c|}
\hline
$\theta_s$ & \begin{tabular}[c]{@{}c@{}}Mean$\pm$ \\ Std. Dev.\end{tabular} & Median & Max. & \begin{tabular}[c]{@{}c@{}}Number of \\ constraints\end{tabular} \\ \hline
0.98       & 3.77$\pm$3.76 &  2.46      & 18.44    &   54                   \\ \hline
0.95       & 2.21$\pm$1.41    &  2.05      & 7.46    &   177                   \\ \hline
0.9       & 1.98$\pm$1.18      &   1.93     & 6.14    &      294                \\ \hline
0.8       &    1.74$\pm$0.89   &    1.69    &   4.05  &     457                 \\ \hline
0.7       &  1.76$\pm$0.83     &  1.78      &  3.49  &  565                    \\ \hline
0.5       &    1.75$\pm$0.95    & 1.83       &  4.25   & 767                      \\ \hline
0.3       &       2.11$\pm$1.12  &  2.03   &  5.41   &    978                  \\ \hline
0.1       &   2.84$\pm$1.32      &    2.89   &   6.81   &       1403               \\ \hline
\end{tabular}
\end{table}

%%%%%%%%%%%%%%%%%%%%%%%%%%%%%%%%%%%%%%%%%%%%%%%%%%%%%%%%%%%%%%%%%%%%%%%%%%%%%%%%%%%%%%%%%%%%%%%%%%%%%%%%
\subsection{Impact of Different Similarity Threshold $\vartheta_s$}
\label{evaluation_rss_threshold}
%%%%%%%%%%%%%%%%%%%%%%%%%%%%%%%%%%%%%%%%%%%%%%%%%%%%%%%%%%%%%%%%%%%%%%%%%%%%%%%%%%%%%%%%%%%%%%%%%%%%%%%%
Next, we examined the accuracy under the impact of different similarity threshold $\vartheta_s$. 
The results are listed in Table \ref{table_sim}. 
We choose a binning size of $b=0.1$ and a RSS threshold of $\vartheta_r=-70$. 
We vary the threshold $\vartheta_s$ from 0.1 to 0.98 to see the accuracy as well as the number of constraints inferred.
As can be seen from Table \ref{table_sim}, 
the threshold has a high impact on the number of constraints, 
thus an impact on the accuracy. 
Applying a high threshold will result in a small number of constraints and a decrease of the accuracy; 
while a small threshold yields a large number of constraints and an improvement of the accuracy. 
For example, we obtain a mean accuracy of 1.74\,m with $\vartheta_s=0.8$, 
which is an improvement of 53.8\% as compared to the accuracy with $\vartheta_s=0.98$ (i.e., 3.77\,m ).
Yet, such an improvement is at the expense of more number of constraints added 
(i.e., 457 constraints with $\vartheta_s=0.8$ as compared to 54 with $\vartheta_s=0.98$).
But the accuracy does not get improved with a threshold smaller than 0.5. 
One reason could be because a low similarity value 
always comes along with a very high covariance,
and has very less strength to correct the odometric error. 
A value of 0.8 seems to be a good trade off between the accuracy and the number of constraints inferred.

%We also show the mean error of the inferred constraints as compared to the ground truth in Fig. XX.
%As can be seen from this figure, a larger threshold gives smaller error of the inferred constraints. 
%The constraints, whose transformations are set be zero, 
%will introduce smaller errors (between xx and xx m) as compared to ground truth, 
%but still allows to compensate for the accumulated error from odometry. 
%From this point of view, 
%fingerprinting-based SLAM differs from the SLAM systems based on visual sensors or laser range finders.

\begin{table}[]
\centering
\caption{accuracy (mean and standard deviation, median, and maximum) in meters under the impact of different binning size $b$.}
\label{table_binning}
\begin{tabular}{|c|c|c|c|c|}
\hline
binning $b$ & Mean$\pm$  Std. Dev.& Median & Maximum \\ \hline
0.05       & 1.78$\pm$0.88       & 1.78       &          3.92             \\ \hline
0.1      & 1.74$\pm$0.89    &  1.69      &         4.05              \\ \hline
0.2       &  2.12$\pm$0.98   &  1.37      &    5.21                  \\ \hline
0.4       &      2.65$\pm$1.39      &2.34        &  6.19                    \\ \hline
1.0       &       4.23$\pm$2.82     &3.15        &   11.76                   \\ \hline
\end{tabular}
\end{table}

%%%%%%%%%%%%%%%%%%%%%%%%%%%%%%%%%%%%%%%%%%%%%%%%%%%%%%%%%%%%%%%%%%%%%%%%%%%%%%%%%%%%%%%%%%%%%%%%%%%%%%%%
\subsection{Impact of the Binning Size of Training}
\label{evaluation_rss_threshold}
%%%%%%%%%%%%%%%%%%%%%%%%%%%%%%%%%%%%%%%%%%%%%%%%%%%%%%%%%%%%%%%%%%%%%%%%%%%%%%%%%%%%%%%%%%%%%%%%%%%%%%%%
In the last series of experiments, 
we examined the accuracy under the impact of various binning size $b$ as shown in Table \ref{table_binning}. 
We choose a similarity threshold of $\vartheta_s=0.8$ and RSS threshold of $\vartheta_r=-70$. 
We set parameter $b$ to the following values to evaluate its impact on the accuracy, 
i.e., $b=\{0.05, 0.1, 0.2, 0.4, 1.0\}$. 
As can be seen from Table \ref{table_binning}, 
the best choice of $b$ is $0.1$. 
The covariance estimated with a large $b$ is usually too large 
to compensate for the error from the odometry, 
while a small $b$ seems to well characterize the model. 
Covariance added here is a key to optimize the pose graph, 
as it is the only information to measure how close 
the two locations are in a loop, 
therefore, a careful examination of the parameter will lead to an improvement of the accuracy.
The covariance is much smaller as compared to the ranging of a Wifi access point (up to 50 meters). 
This is why we are still able to correct the accumulated odometry error. 
The approach presented here provides a way to automatically 
calibrate the uncertainty model using odometric measurements. 

%%%%%%%%%%%%%%%%%%%%%%%%%%%%%%%%%%%%%%%%%%%%%%%%%%%%%%%%%%%%%%%%%%%%%%%%%%%%%%%%%%%%%%%%%%%%%%%%%%%%%%%%
\subsection{Computational Time}
\label{evaluation_rss_threshold}
%%%%%%%%%%%%%%%%%%%%%%%%%%%%%%%%%%%%%%%%%%%%%%%%%%%%%%%%%%%%%%%%%%%%%%%%%%%%%%%%%%%%%%%%%%%%%%%%%%%%%%%%
The time consumed at each stage of our approach is listed in Table \ref{table:computational_time}. 
We processed the measurements with an Intel Core i5-4200M @ 2.5GHz CPU, with 4GB RAM. 
As can be seen from Table \ref{table:computational_time}, the entire data processing of 
radio fingerprint and odometry is almost 70 times faster than the data recording stage. 
Optimization of the graph only took 0.31 seconds.
Although our implementation is offline, 
loop-closure detection and pose graph optimization can be possibly 
made online considering the time consumption of the system as shown in Table \ref{table:computational_time}. 

\begin{table}[]
\centering
\caption{Time consumption (in seconds) at each stage of our approach.}
\label{table:computational_time}
\begin{tabular}{|c|c|}
\hline
Stage                   & Duration (s) \\ \hline
Data recording (time per track) & 815.0        \\ \hline
Model training+variance computation     & 3.49         \\ \hline
Loop closure detection  &    7.35      \\ \hline
Screening of loop candidates      &    0.04           \\ \hline
Pose Graph Optimization &        0.31       \\ \hline
\end{tabular}
\end{table}
%%%%%%%%%%%%%%%%%%%%%%%%%%%%%%%%%%%%%%%%%%%%%%%%%%%%%%%%%%%%%%%%%%%%%%%%%%%%%%%%%%%%%%%%%%%%%%%%%%%%%%%%
\section{\textbf{Conclusions and Future Work}}
\label{conclusions}
%%%%%%%%%%%%%%%%%%%%%%%%%%%%%%%%%%%%%%%%%%%%%%%%%%%%%%%%%%%%%%%%%%%%%%%%%%%%%%%%%%%%%%%%%%%%%%%%%%%%%%%%
In this paper, 
we presented a novel approach for crowd-sensing simultaneous localization and radio fingerprint mapping (C-SLAM-RF) in unknown environments 
using radio signals collected from multiple users. 
The proposed system makes use of an inertial tracking system 
and the signal strength measurements from surrounding wireless access points. 
We evaluated the proposed approach in a large scale environment and an accuracy of 1.74 meters is achieved over an area of approx. 9000 square meters.
%that the accuracy is comparable to a point-cloud-based SLAM system. 
%The error, although, is larger as compared to a typical laser-based positioning system frequently used in robotics, it features a cost-effective solution with .
In the future, we would like to investigate 
how to create a finer radio fingerprint map from the coarsely sampled human trajectories. 
%Another direction will be to relax the orientation assumption we made in the paper. 
Another direction would be to replace Tango with the low-cost IMU sensors embedded in a smartphone 
for pedestrian dead reckoning \cite{smartphone_stepcount,richard_ieeesensor_2016}.
\bibliographystyle{IEEEtran}
\bibliography{literatur}

\begin{thebibliography}{10}
\providecommand{\url}[1]{#1}
\csname url@rmstyle\endcsname
\providecommand{\newblock}{\relax}
\providecommand{\bibinfo}[2]{#2}
\providecommand\BIBentrySTDinterwordspacing{\spaceskip=0pt\relax}
\providecommand\BIBentryALTinterwordstretchfactor{4}
\providecommand\BIBentryALTinterwordspacing{\spaceskip=\fontdimen2\font plus
\BIBentryALTinterwordstretchfactor\fontdimen3\font minus
  \fontdimen4\font\relax}
\providecommand\BIBforeignlanguage[2]{{%
\expandafter\ifx\csname l@#1\endcsname\relax
\typeout{** WARNING: IEEEtran.bst: No hyphenation pattern has been}%
\typeout{** loaded for the language `#1'. Using the pattern for}%
\typeout{** the default language instead.}%
\else
\language=\csname l@#1\endcsname
\fi
#2}}

\bibitem{dissanayake2001solution}
M.~G. Dissanayake, P.~Newman, S.~Clark, H.~F. Durrant-Whyte, and M.~Csorba, ``A
  solution to the simultaneous localization and map building {(SLAM)}
  problem,'' \emph{IEEE Transactions on Robotics and Automation}, vol.~17,
  no.~3, pp. 229--241, 2001.

\bibitem{montemerlo2002fastslam}
M.~Montemerlo, S.~Thrun, D.~Koller, B.~Wegbreit, \emph{et~al.}, ``{FastSLAM}: A
  factored solution to the simultaneous localization and mapping problem,'' in
  \emph{In Proceedings of the AAAI National Conference on Artificial
  Intelligence (AAAI 2002)}, Edmonton, Alberta, Canada, July 28-August 1 2002,
  pp. 593--598.

\bibitem{ferris2007wifi}
B.~Ferris, D.~Fox, and N.~Lawrence, ``{WiFi-SLAM} using gaussian process latent
  variable models,'' in \emph{Proceedings of the 20th International Joint
  Conference on Artifical Intelligence (IJCAI'07)}, Hyderabad, India, January
  06-12 2007, pp. 2480--2485.

\bibitem{Thrun_Probabilistic_robotics}
S.~Thrun, W.~Burgard, and D.~Fox, \emph{Probabilistic Robotics (Intelligent
  Robotics and Autonomous Agents)}.\hskip 1em plus 0.5em minus 0.4em\relax The
  MIT Press, 2005.

\bibitem{kuemmerle11icra}
R.~Kuemmerle, G.~Grisetti, H.~Strasdat, K.~Konolige, and W.~Burgard, ``g2o: A
  general framework for graph optimization,'' in \emph{the 2011 IEEE
  International Conference on Robotics and Automation (ICRA 2011)}, Shanghai,
  China, May 9-13 2011, pp. 3607--3613.

\bibitem{Lu_millos_1997}
F.~Lu and E.~Milios, ``Globally consistent range scan alignment for environment
  mapping,'' \emph{Autonomous Robots}, vol.~4, no.~4, pp. 333--349, October
  1997.

\bibitem{Taketomi2017}
T.~Taketomi, H.~Uchiyama, and S.~Ikeda, ``Visual slam algorithms: a survey from
  2010 to 2016,'' \emph{IPSJ Transactions on Computer Vision and Applications},
  vol.~9, no.~1, p.~16, Jun 2017.

\bibitem{lahiru_access}
L.~{Jayasinghe}, N.~{Wijerathne}, C.~{Yuen}, and M.~{Zhang}, ``Feature learning
  and analysis for cleanliness classification in restrooms,'' \emph{IEEE
  Access}, vol.~7, pp. 14\,871--14\,882, January 2019.

\bibitem{ran_ieee_sensors2017}
R.~Liu, C.~Yuen, T.~N. Do, and U.~X. Tan, ``Fusing similarity-based sequence
  and dead reckoning for indoor positioning without training,'' \emph{IEEE
  Sensors Journal}, vol.~17, no.~13, pp. 4197--4207, July 2017.

\bibitem{billy_iot}
B.~P.~L. {Lau}, N.~{Wijerathne}, B.~K.~K. {Ng}, and C.~{Yuen}, ``Sensor fusion
  for public space utilization monitoring in a smart city,'' \emph{IEEE
  Internet of Things Journal}, vol.~5, no.~2, pp. 473--481, April 2018.

\bibitem{ran_localize_AP}
R.~{Liu}, M.~{Padmal}, S.~H. {Marakkalage}, T.~{Shaganan}, C.~{Yuen}, and
  U.~{Tan}, ``Localizing heterogeneous access points using similarity-based
  sequence,'' in \emph{2018 3rd International Conference on Advanced Robotics
  and Mechatronics (ICARM)}, July 2018, pp. 306--311.

\bibitem{liuran_spawc}
R.~{Liu}, C.~{Yuen}, T.~{Do}, Y.~{Jiang}, X.~{Liu}, and U.~{Tan}, ``Indoor
  positioning using similarity-based sequence and dead reckoning without
  training,'' in \emph{2017 IEEE 18th International Workshop on Signal
  Processing Advances in Wireless Communications (SPAWC)}, July 2017, pp. 1--5.

\bibitem{Huang_wifi_slam_11}
J.~Huang, D.~Millman, M.~Quigley, D.~Stavens, S.~Thrun, and A.~Aggarwal,
  ``Efficient, generalized indoor wifi graphslam,'' in \emph{2011 IEEE
  International Conference on Robotics and Automation (ICRA 2011)}, Shanghai,
  China, May 09-13 2011, pp. 1038--1043.

\bibitem{Yassin_ieee_tutorials_2016}
A.~Yassin, Y.~Nasser, M.~Awad, A.~Al-Dubai, R.~Liu, C.~Yuen, and R.~Raulefs,
  ``Recent advances in indoor localization: A survey on theoretical approaches
  and applications,'' \emph{IEEE Communications Surveys and Tutorials},
  vol.~19, no.~2, pp. 1327--1346, Secondquarter, 2017.

\bibitem{he2016wi}
S.~He and S.-H.~G. Chan, ``Wi-fi fingerprint-based indoor positioning: Recent
  advances and comparisons,'' \emph{IEEE Communications Surveys and Tutorials},
  vol.~18, no.~1, pp. 466--490, 2016.

\bibitem{durrant2006simultaneous}
H.~Durrant-Whyte and T.~Bailey, ``Simultaneous localization and mapping: part
  i,'' \emph{IEEE robotics \& automation magazine}, vol.~13, no.~2, pp.
  99--110, 2006.

\bibitem{bailey2006simultaneous}
T.~Bailey and H.~Durrant-Whyte, ``Simultaneous localization and mapping (slam):
  Part ii,'' \emph{IEEE Robotics \& Automation Magazine}, vol.~13, no.~3, pp.
  108--117, 2006.

\bibitem{engelhard2011real}
N.~Engelhard, F.~Endres, J.~Hess, J.~Sturm, and W.~Burgard, ``Real-time 3d
  visual slam with a hand-held rgb-d camera,'' in \emph{Proc. of the RGB-D
  Workshop on 3D Perception in Robotics at the European Robotics Forum,
  Vasteras, Sweden}, vol. 180, 2011, pp. 1--15.

\bibitem{kim2007slam}
H.-D. Kim, S.-W. Seo, I.-h. Jang, and K.-B. Sim, ``Slam of mobile robot in the
  indoor environment with digital magnetic compass and ultrasonic sensors,'' in
  \emph{International Conference on Control, Automation and Systems
  (ICCAS'07)}, Seoul, South Korea, October 17-20 2007, pp. 87--90.

\bibitem{vallivaara2011magnetic}
I.~Vallivaara, J.~Haverinen, A.~Kemppainen, and J.~R{\"o}ning, ``Magnetic
  field-based slam method for solving the localization problem in mobile robot
  floor-cleaning task,'' in \emph{the 15th International Conference on Advanced
  Robotics (ICAR 2011)}, Tallinn, Estonia, June 20-23 2011, pp. 198--203.

\bibitem{jirkuu2016wifi}
M.~Jirku, V.~Kubelka, and M.~Reinstein, ``Wifi localization in 3d,'' in
  \emph{Proceedings of the 2016 IEEE/RSJ International Conference on
  Intelligent Robots and Systems (IROS 2016)}, Daejeon, Korea, October 9-14
  2016, pp. 4551--4557.

\bibitem{signal_slam}
P.~Mirowski, T.~K. Ho, S.~Yi, and M.~MacDonald, ``Signalslam: Simultaneous
  localization and mapping with mixed wifi, bluetooth, lte and magnetic
  signals,'' in \emph{International Conference on Indoor Positioning and Indoor
  Navigation}, Montbeliard-Belfort, France, October 28-31 2013, pp. 1--10.

\bibitem{yang2012locating}
Z.~Yang, C.~Wu, and Y.~Liu, ``Locating in fingerprint space: wireless indoor
  localization with little human intervention,'' in \emph{Proceedings of the
  18th Annual International Conference on Mobile Computing and Networking},
  Istanbul, Turkey, August 22-26 2012, pp. 269--280.

\bibitem{subbu2014analysis}
K.~Subbu, C.~Zhang, J.~Luo, and A.~Vasilakos, ``Analysis and status quo of
  smartphone-based indoor localization systems,'' \emph{IEEE Wireless
  Communications}, vol.~21, no.~4, pp. 106--112, 2014.

\bibitem{dong2017unleashing}
J.~Dong, ``Unleashing the power of the crowd: Towards efficient and sustainable
  mobile crowdsensing,'' Ph.D. dissertation, 2017.

\bibitem{marakkalage2018understanding}
S.~H. {Marakkalage}, S.~{Sarica}, B.~P.~L. {Lau}, S.~K. {Viswanath},
  T.~{Balasubramaniam}, C.~{Yuen}, B.~{Yuen}, J.~{Luo}, and R.~{Nayak},
  ``Understanding the lifestyle of older population: Mobile crowdsensing
  approach,'' \emph{IEEE Transactions on Computational Social Systems}, vol.~6,
  no.~1, pp. 82--95, Feb 2019.

\bibitem{faragher2012opportunistic}
R.~Faragher, C.~Sarno, and M.~Newman, ``Opportunistic radio slam for indoor
  navigation using smartphone sensors,'' in \emph{Proceedings of the 2012
  IEEE/ION Position Location and Navigation Symposium (PLANS 2012)}, Myrtle
  Beach, South Carolina, USA, April 23-26 2012, pp. 120--128.

\bibitem{gao2014jigsaw}
R.~Gao, M.~Zhao, T.~Ye, F.~Ye, Y.~Wang, K.~Bian, T.~Wang, and X.~Li, ``Jigsaw:
  Indoor floor plan reconstruction via mobile crowdsensing,'' in
  \emph{Proceedings of the 20th Annual International Conference on Mobile
  Computing and Networking}, Maui, Hawaii, USA, September 07-11 2014, pp.
  249--260.

\bibitem{radu2013pazl}
V.~Radu, L.~Kriara, and M.~K. Marina, ``Pazl: A mobile crowdsensing based
  indoor wifi monitoring system,'' in \emph{Proceedings of the 9th
  International Conference on Network and Service Management (CNSM 2013)},
  Z\"urich, Switzerland, 2013, pp. 75--83.

\bibitem{RFM-SLAM}
V.~S. Saurav~Agarwal and S.~Chakravorty, ``Rfm-slam: Exploiting relative
  feature measurements to separate orientation and position estimation in
  slam,'' in \emph{In Proc. IEEE International Conference on Robotics and
  Automation (ICRA)}, Singapore, May 29--June 3 2017.

\bibitem{slam_trends_2016}
C.~Cadena, L.~Carlone, H.~Carrillo, Y.~Latif, D.~Scaramuzza, J.~Neira, I.~Reid,
  and J.~J. Leonard, ``Past, present, and future of simultaneous localization
  and mapping: Toward the robust-perception age,'' \emph{IEEE Transactions on
  Robotics}, vol.~32, no.~6, pp. 1309--1332, Dec 2016.

\bibitem{ceres-solver}
S.~Agarwal, K.~Mierle, and Others, ``Ceres solver,''
  \url{http://ceres-solver.org}.

\bibitem{SujiwoATNE16}
A.~Sujiwo, T.~Ando, E.~Takeuchi, Y.~Ninomiya, and M.~Edahiro, ``Monocular
  vision-based localization using {ORB-SLAM} with lidar-aided mapping in
  real-world robot challenge,'' \emph{{Journal of Robotics and Mechatronics}},
  vol.~28, no.~4, pp. 479--490, 2016.

\bibitem{gao2018ldso}
X.~Gao, R.~Wang, N.~Demmel, and D.~Cremers, ``Ldso: Direct sparse odometry with
  loop closure,'' in \emph{2018 IEEE/RSJ International Conference on
  Intelligent Robots and Systems (IROS 2018)}, Madrid, Spain, October 1-5 2018.

\bibitem{RanArtur_IROS_2012}
R.~Liu, A.~Koch, and A.~Zell, ``Path following with passive {UHF RFID} received
  signal strength in unknown environments,'' in \emph{Proc. of the 2012
  IEEE/RSJ Int. Conf. on Intelligent Robots and Systems (IROS 2012)},
  Vilamoura, Algarve, Portugal, October 2012, pp. 2250--2255.

\bibitem{vorst2010isr}
P.~Vorst and A.~Zell, ``A comparison of similarity measures for localization
  with passive {RFID} fingerprints,'' in \emph{the Joint Conference of 41st
  International Symposium on Robotics and 6th German Conference on Robotics},
  Munich, Germany, June 2010, pp. 354--361.

\bibitem{vorst2011rfidta}
P.~Vorst, A.~Koch, and A.~Zell, ``Efficient self-adjusting, similarity-based
  location fingerprinting with passive {UHF} {RFID},'' in \emph{the IEEE
  International Conference on {RFID}-Technology and Applications (RFID-TA
  2011)}, Spain, September 2011, pp. 160--167.

\bibitem{pdr_arbitrary_placement}
Z.~Xiao, H.~Wen, A.~Markham, and N.~Trigoni, ``Robust pedestrian dead reckoning
  {(R-PDR)} for arbitrary mobile device placement,'' in \emph{the 2014
  International Conference on Indoor Positioning and Indoor Navigation}, Busan,
  Korea, October 27--30 2014, pp. 187--196.

\bibitem{GalvezIROS11}
D.~Galvez-Lopez and J.~D. Tardos, ``Real-time loop detection with bags of
  binary words,'' in \emph{2011 IEEE/RSJ International Conference on
  Intelligent Robots and Systems (IROS 2011)}, San Francisco, CA, USA,
  September 25-30 2011, pp. 51--58.

\bibitem{pcl_icra_2011}
R.~B. Rusu and S.~Cousins, ``3d is here: Point cloud library (pcl),'' in
  \emph{2011 IEEE International Conference on Robotics and Automation (ICRA
  2011)}, Shanghai, China, May 09-13 2011, pp. 1--4.

\bibitem{Alexandre123ddescriptors}
L.~A. Alexandre, ``3d descriptors for object and category recognition: a
  comparative evaluation,'' in \emph{Workshop on Color-Depth Camera Fusion in
  Robotics at the IEEE/RSJ International Conference on Intelligent Robots and
  Systems (IROS)}, Vilamoura, Algarve Portugal, October 07-12 2012.

\bibitem{smartphone_stepcount}
F.~Gu, K.~Khoshelham, J.~Shang, F.~Yu, and Z.~Wei, ``Robust and accurate
  smartphone-based step counting for indoor localization,'' \emph{IEEE Sensors
  Journal}, vol.~17, no.~11, pp. 3453--3460, June 2017.

\bibitem{richard_ieeesensor_2016}
T.~N. Do, R.~Liu, C.~Yuen, M.~Zhang, and {U-X. Tan}, ``Personal dead reckoning
  using imu mounted on upper torso and inverted pendulum model,'' \emph{IEEE
  Sensors Journal}, vol.~16, no.~21, pp. 7600--7608, November 1 2016.

\end{thebibliography}

\end{document}